\documentclass[letterpaper]{article}
\usepackage[preprint]{aaai2027}

\usepackage[hyphens]{url} 
\usepackage{graphicx} 
\usepackage{natbib} 
\usepackage{caption} 
\usepackage{amsmath}
\usepackage{amssymb}
\usepackage{bm}
\usepackage{booktabs}
\usepackage{multirow}
\usepackage{xcolor}
\usepackage{colortbl}
\usepackage{tikz}
\usepackage{enumitem}

\renewcommand{\paragraph}[1]{\smallskip\noindent\textbf{#1.}}
\renewcommand{\subparagraph}[1]{\smallskip\noindent\textbf{\underline{#1.}}}

\begin{document}

\title{IDRBench: Benchmarking the Interactive Capabilities of Deep Research Agents}

\author{
    Yingchaojie Feng\textsuperscript{\rm 1},
    Qiang Huang\textsuperscript{\rm 2}\corresponding,
    Xiaoya Xie\textsuperscript{\rm 3},
    Zhaorui Yang\textsuperscript{\rm 4},\\
    Jun Yu\textsuperscript{\rm 2},
    Wei Chen\textsuperscript{\rm 4},
    Anthony K. H. Tung\textsuperscript{\rm 1}
}
\affiliations{
    \textsuperscript{\rm 1}School of Computing, National University of Singapore\\
    \textsuperscript{\rm 2}School of Intelligence Science and Engineering, Harbin Institute of Technology (Shenzhen)\\
    \textsuperscript{\rm 3}Zhejiang University\\
    \textsuperscript{\rm 4}State Key Lab of CAD\&CG, Zhejiang University
}

\maketitle

\begin{abstract}
Large Language Model (LLM)-based deep research agents perform multi-step reasoning, web exploration, and long-form report generation.
In these long-horizon workflows, early deviations from user intent can misdirect research and propagate through planning, search, and synthesis, making timely interaction essential.
However, existing benchmarks primarily treat deep research as a static input-output task, overlooking agents' ability to elicit and use user feedback.
We introduce \textbf{IDRBench}, a benchmark for evaluating interactive deep research with controlled opportunities for clarification.
Within a common workflow and stage-wise interaction budget, IDRBench compares autonomous and interactive trajectories, measuring interaction benefit through changes in task-specific report alignment and interaction cost through turns and tokens.
Comprehensive experiments on 100 tasks with seven proprietary and open-weight LLMs show that interaction improves all five alignment measures for every model, yielding an average gain of 6.39 points, while revealing distinct trade-offs among autonomous performance, alignment gain, and communication cost.
At the task level, interaction improves performance in 74.4\% of cases but degrades it in 19.9\%, demonstrating that access to clarification alone does not guarantee better outcomes: success depends on what agents ask and how effectively they incorporate the resulting feedback.
\end{abstract}

\section{Introduction}
\label{sec:intro}

Large Language Models (LLMs) have revolutionized information seeking, evolving from single-turn question answering to deep research agents that independently reason, navigate the web, and generate long-form reports \cite{zheng2024openresearcher, li2025search, zheng2025deepresearcher, yun2025interaction}. 
Unlike traditional Retrieval-Augmented Generation (RAG) systems \cite{lewis2020retrieval, gao2023retrieval, wang2024searching} that handle isolated queries, these agents iteratively plan, search, and synthesize information to address open-ended user needs \cite{wei2025browsecomp, du2026deepresearch}.

\begin{figure}[t]
  \centering
  \includegraphics[width=0.99\columnwidth]{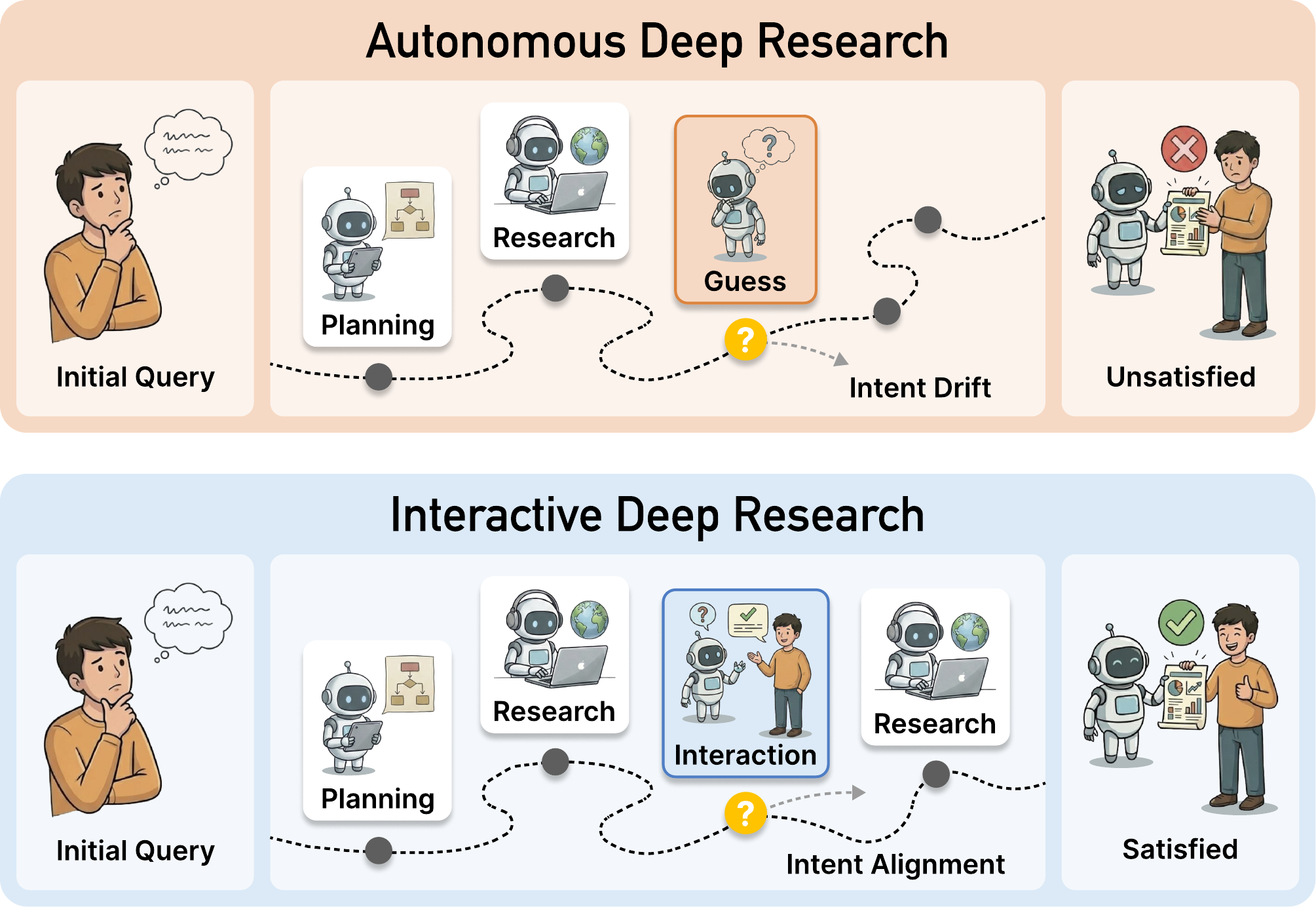}
  \caption{\textbf{Comparison of autonomous and interactive deep research.} Autonomous agents execute independently, while interactive agents can request feedback to guide research.}
  \label{fig:teaser-example}
\end{figure}

Despite these advances, deep research remains largely \emph{autonomous}, with agents controlling the trajectory after the initial query \cite{zheng2025deepresearcher}.
As new evidence and alternatives emerge, choices about scope, emphasis, audience, and time horizon can amplify early intent mismatches throughout planning, search, and synthesis.
Recent clarification mechanisms for planning and tool use \cite{zhang2024ask, zhang2025asktoact, zhang2024clamber} motivate interactive deep research, where agents solicit feedback to maintain alignment (\textbf{Figure~\ref{fig:teaser-example}}).

Yet, benchmark design has not kept pace with this shift. They largely formulate deep research as a static input-output task and evaluate final reports through retrieval, factuality, and citation-based metrics, leaving three important gaps:
\begin{itemize}[nolistsep]
  \item \textbf{Autonomous Task Formulation:}
  Agents receive fixed prompts and complete the entire research process without opportunities to seek clarification or refine user intent \cite{du2026deepresearch, coelho2026deepresearchgym, zhou2025browsecomp}.

  \item \textbf{Lack of Controlled Interaction Environment:}
  Reliable comparison requires consistent feedback, while repeated human-in-the-loop evaluation is costly and difficult to scale \cite{yoon2024usersim, ye2025productagent, yao2025taubench}.

  \item \textbf{Outcome-Only Evaluation:}
  By focusing on final reports, existing benchmarks overlook whether interaction improves intent alignment and the cost it introduces during research \cite{shao2024assisting, wu2025writingbench, du2026deepresearch}.
\end{itemize}

To bridge these gaps, we introduce \textbf{IDRBench}, an \textbf{I}nteractive \textbf{D}eep \textbf{R}esearch \textbf{Bench}mark for systematically evaluating the interactive capabilities of deep research agents.
Its benchmark protocol compares complete autonomous and interactive trajectories under the same research workflow, tools, execution limits, and task target. In the interactive setting, models receive a shared stage-wise clarification budget and standardized, question-conditioned feedback, while independently deciding whether, when, and what to ask and how to use the response.
IDRBench constructs tasks with varying levels of query specificity while retaining the corresponding reports as consistent evaluation targets.
Its interaction-aware metrics jointly capture outcome and burden: Report Alignment measures document-, structure-, and aspect-level agreement with the task target, while Interaction Cost quantifies the associated turns and tokens.

We evaluate seven representative proprietary and open-weight LLMs across 100 research tasks.
Interaction improves all five alignment measures for every model, yielding an average gain of 6.39 points.
However, the models exhibit distinct interactive profiles: GPT-5.1 achieves the highest autonomous score (75.59), Claude-Sonnet-4.5 the highest interactive score (80.47), and Llama-4-Maverick the largest improvement (+10.96).
At the task level, interaction improves performance in 74.4\% of cases but degrades it in 19.9\%, demonstrating that clarification opportunities alone do not guarantee better outcomes. Additional diagnostics across feedback models and context settings further characterize the stability of the evaluation protocol.
These findings establish interactive capability as a distinct dimension complementary to autonomous model strength and position IDRBench as a reusable testbed for controlled comparison.

Our contributions are threefold:
\begin{itemize}
  \item \textbf{Interactive Deep Research Benchmark:} We introduce IDRBench, which extends deep research evaluation from static input-output assessment to complete autonomous and interactive research trajectories.

  \item \textbf{Controlled Interaction Protocol:} We develop a controlled protocol that provides consistent, task-grounded responses under a shared stage-wise clarification budget, enabling scalable and reproducible comparison without repeated human-in-the-loop evaluation.

  \item \textbf{Interaction-Aware Metrics:} We jointly measure task-specific report-alignment and interaction cost, revealing how models differ in when they seek clarification, what information they elicit, and how effectively they use the resulting feedback.
\end{itemize}

\section{Related Work}
\label{sec:related_work}

\paragraph{Deep Research Frameworks}
Deep research systems combine long-horizon reasoning, web exploration, and citation-grounded reporting through iterative planning and tool use~\cite{shao2024assisting, coelho2026deepresearchgym, guo2025deepseek, zhou2024trustworthiness, zhao2026retrieval}.
Existing approaches include multi-agent frameworks with specialized roles~\cite{zheng2024openresearcher, alzubi2025open, li2025search} and end-to-end agents trained with reinforcement learning~\cite{jin2025search, zheng2025deepresearcher}.
Despite their growing capabilities, these systems are predominantly designed and evaluated for autonomous execution, providing limited insight into how agents seek and use user guidance. 
Rather than proposing another agent architecture, IDRBench instantiates a common deep research workflow with controlled clarification opportunities across planning, research, and generation, enabling systematic comparison of interactive behavior.

\paragraph{Deep Research Benchmarks}
Existing benchmarks evaluate individual capabilities such as web retrieval \cite{wei2025browsecomp, zhou2025browsecomp}, long-form writing \cite{bai2024longwriter, wu2025writingbench}, citation-grounded reports \cite{shao2024assisting}, and long-horizon research \cite{du2026deepresearch}.
However, most formulate evaluation as a fixed prompt-to-report task and assess only the final output. Consequently, they cannot determine whether research-time clarification improves alignment with user intent or quantify the communication overhead it introduces.
IDRBench addresses this limitation by comparing complete autonomous and interactive trajectories for the same model and task under matched tools, execution limits, and evaluation criteria.

\paragraph{Interactive Agents}
Clarification has been extensively studied in conversational search for resolving underspecified information needs \cite{rahmani2023survey, tavakoli2022mimics, feng2023towards, aliannejadi2021building}.
Recent LLM agents extend this capability to planning and tool use~\cite{zhang2024ask, zhang2025asktoact, zhang2024clamber}, while $\tau$-bench~\cite{yao2025taubench} and Ask-or-Assume~\cite{edwards2026ask} employ simulated users for scalable evaluation.
Within deep research, Interaction-Driven Browsing~\cite{yun2025interaction} and STEER~\cite{ai2026an} incorporate iterative feedback, with STEER emphasizing adaptive pausing and live persona modeling. 
These studies demonstrate the value of interaction but do not provide a common protocol for comparing existing deep research agents across complete autonomous and interactive trajectories.
Drawing on controlled-feedback environments used in conversational recommendation, product search, tool use, and social-agent evaluation \cite{yoon2024usersim, ye2025productagent, lu2025toolsandbox, zhou2024sotopia}, IDRBench provides consistent, question-conditioned feedback under a shared stage-wise budget.

\section{IDRBench}
\label{sec:method}

\begin{figure*}[!t]
  \centering
  \includegraphics[width=0.99\textwidth]{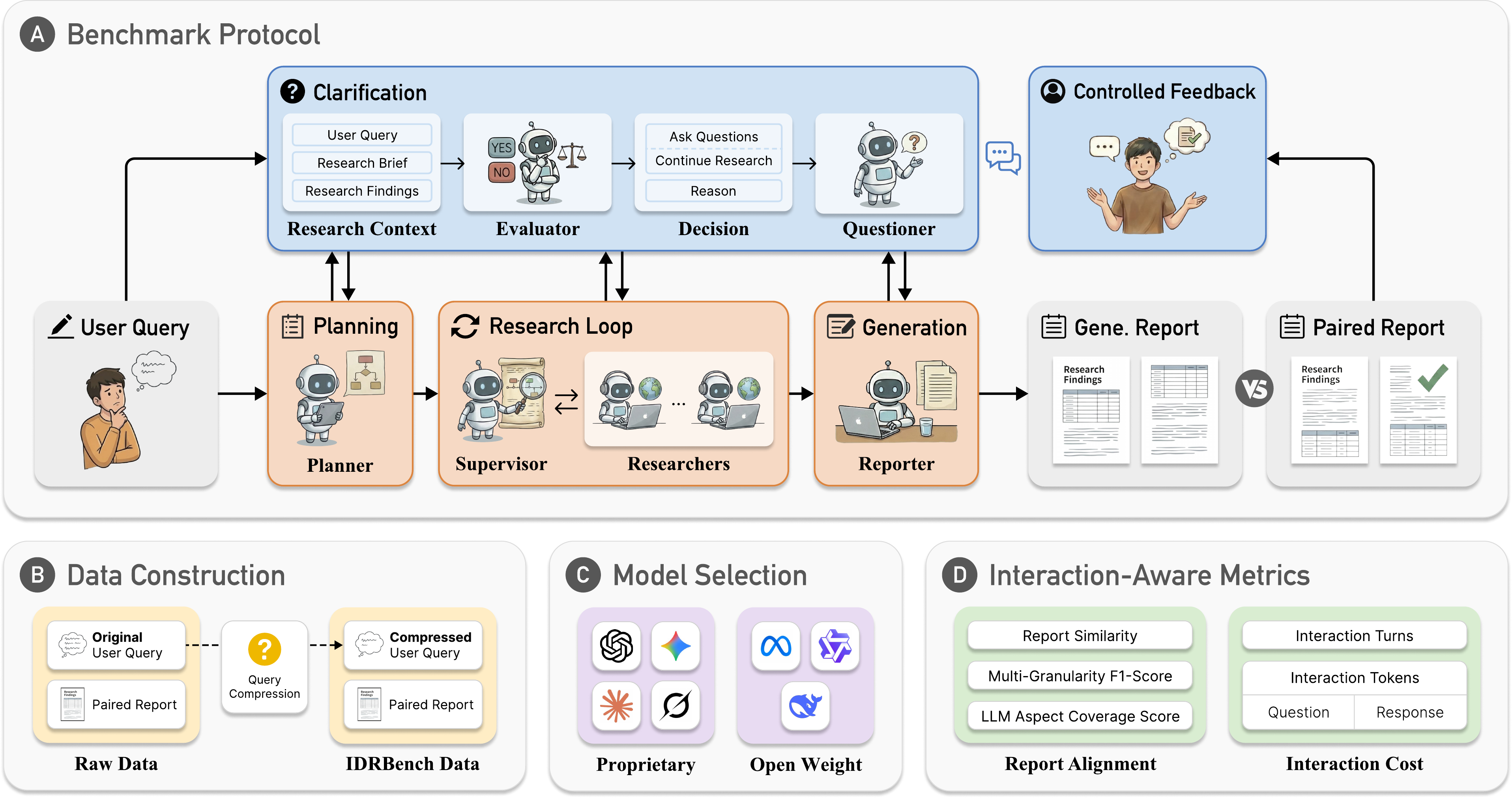}
  \caption{\textbf{Overview of IDRBench.} The benchmark combines a common research workflow with controlled clarification, query compression, model selection, and evaluation of task-specific report alignment and interaction cost.}
  \label{fig:framework}
\end{figure*}

We present \textbf{IDRBench}, a benchmark for systematically comparing deep research agents under both autonomous and interactive task settings (\textbf{Figure \ref{fig:framework}}).

\subsection{Benchmark Protocol}
\label{sec:method:framework}

\paragraph{Task Setting}
Open-ended research requests can support multiple reasonable reports, and users may not exhaustively specify task-specific preferences, such as coverage and emphasis, in the initial query. 
During interaction, users can express these preferences in response to an agent's questions, guiding subsequent planning, search, and generation. 
The same preferences also inform how users assess whether the completed report meets their expectations. 
IDRBench follows this setting by pairing a broader query with a report that reflects one set of task-specific choices. 
The paired report provides a consistent task target for feedback and evaluation across models. It represents one reasonable outcome for the query among other possible reports.

For each model-task pair, IDRBench compares autonomous and interactive trajectories under the same research backbone, retrieval tools, execution limits, evaluation metrics, and task target. 
The interactive condition provides the same stage-wise clarification budget to every model. 
The feedback available in each trajectory depends on the questions the agent asks, while the final report also depends on how the replies are used throughout the remaining research process. 
IDRBench therefore compares the task-specific report-alignment change realized by each model within the common clarification budget, together with the interaction turns and tokens it uses.

\paragraph{Basic Architecture}
To support this comparison, we instantiate a common workflow using the \texttt{langchain-ai} open deep research architecture~\cite{langchain-ai2025open}, a modular pipeline comprising Planning, Research Loop, and Generation. These stages expose comparable interaction points while keeping execution conditions consistent across models.
The pipeline coordinates four agents. 
The Planner converts the query into a structured research brief. During research, the Supervisor decomposes the brief, assigns sub-tasks to Researchers, and monitors progress. Researchers explore assigned topics and summarize findings, while the Reporter synthesizes them into a coherent long-form report.

\paragraph{Controlled Interaction Protocol}
IDRBench permits bounded interaction during Planning, Research Loop, and Generation. 
Every model in the interactive setting receives stage-wise maxima of one, three, and one clarification turn. 
Within this budget, \emph{Clarification} determines whether, when, and what to ask, while \emph{Controlled Feedback} returns a reply under a standardized policy.
The stages cover initial framing, evidence-driven adjustment, and final emphasis, respectively.

\subparagraph{Clarification}
Clarification involves two roles. The Evaluator determines whether interaction is warranted by balancing the benefit of resolving ambiguity against the interruption burden, based on the current research context and remaining budget. 
When interaction is triggered, the Questioner uses the Evaluator's rationale and relevant research state to formulate one or two focused questions about scope, direction, or emphasis, adapting to the user's language style. 
The two-role design supports controlled stage-wise comparisons (Section~\ref{sec:expt:interaction-timing}) by triggering question generation at designated stages without an Evaluator decision.

\subparagraph{Controlled Feedback}
Controlled Feedback provides question-conditioned guidance under a common policy across models, while keeping private task context inaccessible to the research agent. 
The guidance available in each trajectory therefore depends on what the agent asks. 
We implement this role with an LLM policy that answers clarification questions in a natural first-person style using the paired report as private task context. The policy is instructed to remain question-relevant and high-level, without directly exposing the report or introducing unsupported requirements.
Prompt details are provided in \textbf{Appendix A}.

\subsection{Data Construction}
\label{sec:method:data}

IDRBench uses 100 DeepResearch Bench query-report pairs spanning 22 fields and evenly split between Chinese and English~\cite{du2026deepresearch}. This scale is consistent with costly, long-horizon agent evaluations that involve multi-step planning, web exploration, and long-form generation.
However, source queries are often highly detailed (up to $\sim$800 tokens), leaving limited room for meaningful clarification.

We therefore use an LLM to compress each query by 10\%--90\%, varying its specificity while retaining its topic and core objective (see \textbf{Appendix B}).
Each compressed query remains paired with its original report as the task target.
We manually verify all queries for consistency, naturalness, and safety.
Ethical considerations are provided in \textbf{Appendix C}.

\subsection{Model Selection}
\label{sec:method:model}

To cover diverse modeling paradigms, we study four proprietary LLMs: GPT-5.1 \cite{openai2025gpt51}, Gemini-2.5-Pro \cite{comanici2025gemini}, Claude-Sonnet-4.5 \cite{anthropic2025sonnet45}, and Grok-4.1-Fast \cite{xai2025grok41}. 
These represent leading commercial systems optimized for long-context reasoning and tool use. 
We also include three open-weight models, Qwen3-235B \cite{yang2025qwen3}, Llama-4-Maverick \cite{meta2025llama4}, and DeepSeek-V3.2 \cite{liu2025deepseek}, spanning varied scaling and alignment characteristics.

Although Gemini-3-series Pro models \cite{deepmind2026gemini31pro} are newer, preliminary experiments showed unstable compatibility with the LangChain-based long-horizon workflow. We therefore use Gemini-2.5-Pro for more reliable structured prompting and tool invocation.

\subsection{Interaction-Aware Metrics}
\label{sec:method:evaluation}

Because clarification can redirect the entire research trajectory, IDRBench evaluates each model's joint policy for whether, when, and what to ask and how to use feedback, rather than attributing effects to individual questions.
It captures both outcome and burden through Report Alignment, which measures agreement with the task target, and Interaction Cost, which measures turns and tokens.

\paragraph{Report Alignment}
Report Alignment is evaluated through three complementary views: document-level semantic similarity, multi-granularity structural coverage, and aspect-level coverage of task requirements. 
These measures characterize task-specific coverage and emphasis and complement evaluations of factual correctness, citation reliability, source grounding, and other general report-quality criteria.

\subparagraph{Report Similarity}
Report Similarity measures global semantic alignment between the generated and paired reports beyond exact wording.
Let $\bm{e}(\cdot) \in \mathbb{R}^{d}$ denote a text embedding function. We compute normalized cosine similarity as:
\begin{equation}\label{eqn:report-sim}
  \mathrm{sim}(D^{\text{ref}}, D^{\text{gen}}) = \textstyle \frac{1+\cos(\bm{e}(D^{\text{ref}}), \bm{e}(D^{\text{gen}}))}{2}.
\end{equation}
We instantiate $\bm{e}(\cdot)$ with \texttt{Qwen3-Embedding-0.6B}, whose 32K context covers all evaluated reports.
Higher values indicate closer document-level semantic alignment.

\subparagraph{Multi-Granularity F1-Score}
To assess semantic coverage at finer structural units, we compute F1-scores at sentence, paragraph, and chunk levels.
Let $\mathcal{U}^{\text{ref}}=\{\bm{u}_k\}_{k=1}^{K}$ and $\mathcal{U}^{\text{gen}}=\{\bm{v}_i\}_{i=1}^{N}$.
Recall ($R$) and Precision ($P$) are defined as:
\begin{align}
  R &= \textstyle \frac{1}{K}\sum_{k=1}^{K} \bm{1}\bigl[\max_{i}\,\mathrm{sim}(\bm{u}_k, \bm{v}_i)\ge\tau\bigr], \label{eqn:recall} \\
  P &= \textstyle \frac{1}{N}\sum_{i=1}^{N} \bm{1}\bigl[\max_{k}\,\mathrm{sim}(\bm{v}_i, \bm{u}_k)\ge\tau\bigr], \label{eqn:precision}
\end{align}
where $\tau=0.8$.
For chunk-level evaluation, documents are segmented into 300-token windows with a 50-token overlap.
The harmonic-mean F1-Score captures both coverage of the paired report (recall) and whether generated content has a sufficiently similar counterpart in that report (precision).

\subparagraph{LLM Aspect Coverage Score (LLM-ACS)}
It evaluates how well a generated report covers requested task aspects.
Given the detailed source query, we generate $M \in [8,20]$ aspects $\{a_j\}_{1\leq j\leq M}$, each representing a requested informational facet.
For each aspect, let $g_j^{\text{ref}}$ and $g_j^{\text{gen}}$ denote coverage scores for the paired and generated reports, respectively.
Let $\epsilon=10^{-9}$. The final score is computed as:
\begin{equation}\label{eqn:llm-acs}
  \mathrm{LLM\text{-}ACS} = \textstyle \frac{1}{M}\sum_{j=1}^{M} \mathrm{clip}\bigl(\tfrac{g_j^{\text{gen}}}{g_j^{\text{ref}}+\epsilon},0,1\bigr).
\end{equation}
We obtain the 0--5 coverage scores using GPT-4.1 (temperature 0).
Across 20,202 scored aspects, $g_j^{\text{ref}}\in[1,5]$, avoiding zero denominators.
The Average Score is an unweighted descriptive mean of the five metrics, not a calibrated utility measure or independent replication.

\paragraph{Interaction Cost}
Beyond Report Alignment, effective interaction balances alignment gains against communication costs, measured in Interaction Turns and Interaction Tokens.

\subparagraph{Interaction Turns}
Interaction turns measure how often agents request feedback and allocate their shared stage-wise budget across Planning, Research Loop, and Generation.
Because every model operates under the same stage-wise limits, turn counts provide a comparable view of interaction frequency and budget allocation. 
More turns may improve clarification but also increase interruptions and latency.

\subparagraph{Interaction Tokens}
Interaction tokens quantify communication per interaction.
We separately report \emph{question tokens} sent by the agent and \emph{response tokens} returned by the feedback provider. 
Token usage reflects the trade-off between informativeness and communication cost; therefore, fewer tokens are not necessarily better.

\section{Experiments}
\label{sec:expt}

\begin{table*}[t]
\centering
\small
\renewcommand{\arraystretch}{1.0}
\resizebox{\textwidth}{!}{%
\begin{tabular}{llccccccc}
\toprule
\multirow{2}{*}[-0.5ex]{\textbf{Model}} &
\multirow{2}{*}[-0.5ex]{\textbf{Mode}} &
\multirow{2}{*}[-0.75ex]{\textbf{\shortstack{Report \\ Similarity $\uparrow$}}} &
\multicolumn{3}{c}{\textbf{F1-Score $\uparrow$}} &
\multirow{2}{*}[-0.5ex]{\textbf{LLM-ACS $\uparrow$}} &
\multirow{2}{*}[-0.75ex]{\textbf{\shortstack{Avg. Align. \\ Score $\uparrow$}}} &
\multirow{2}{*}[-0.75ex]{\textbf{\shortstack{Est. API Cost \\ (\$/Report) $\downarrow$}}} \\
\cmidrule(lr){4-6}
 & & & \textbf{Sentence} & \textbf{Paragraph} & \textbf{Chunk} & & & \\
\midrule
\multirow{3}{*}{\textbf{GPT-5.1}}
  & \textbf{Autonomous} & 84.92 & \underline{46.05} & 69.07 & \underline{82.30} & 95.61 & \underline{75.59} & 0.473 \\
  & \textbf{Interactive} & 87.54 & 50.44 & 71.99 & 88.08 & 96.79 & 78.97 & 0.586 \\
  & \textbf{Interact. Gain} & +2.61 & +4.39 & +2.91 & +5.77 & +1.18 & +3.37 & +0.113 \\
\midrule

\multirow{3}{*}{\textbf{Gemini-2.5-Pro}}
  & \textbf{Autonomous} & 85.00 & 38.36 & \underline{76.62} & 80.92 & 86.37 & 73.45 & 0.393 \\
  & \textbf{Interactive} & 88.88 & 46.60 & \textbf{82.15} & \textbf{89.21} & 92.60 & 79.89 & 0.752 \\
  & \textbf{Interact. Gain} & +3.88 & +8.25 & +5.52 & +8.29 & +6.22 & +6.43 & +0.359 \\
\midrule

\multirow{3}{*}{\textbf{Claude-Sonnet-4.5}}
  & \textbf{Autonomous} & \underline{85.96} & 44.98 & 69.20 & 81.52 & \underline{95.88} & 75.51 & 0.987 \\
  & \textbf{Interactive} & \textbf{89.15} & \textbf{52.92} & 74.20 & 88.06 & \textbf{98.00} & \textbf{80.47} & 2.220 \\
  & \textbf{Interact. Gain} & +3.19 & +7.94 & +5.00 & +6.54 & +2.13 & +4.96 & +1.233 \\
\midrule

\multirow{3}{*}{\textbf{Grok-4.1-Fast}}
  & \textbf{Autonomous} & 81.28 & 30.76 & 65.33 & 72.93 & 87.44 & 67.55 & 0.192 \\
  & \textbf{Interactive} & 86.68 & 38.63 & 76.47 & 83.24 & 92.56 & 75.52 & 0.275 \\
  & \textbf{Interact. Gain} & +5.41 & +7.87 & +11.14 & +10.31 & +5.12 & +7.97 & +0.083 \\
\midrule

\multirow{3}{*}{\textbf{Llama-4-Maverick}}
  & \textbf{Autonomous} & 76.06 & 18.44 & 64.72 & 61.78 & 53.06 & 54.81 & \underline{0.021} \\
  & \textbf{Interactive} & 83.93 & 24.65 & 78.46 & 75.31 & 66.53 & 65.78 & \textbf{0.026} \\
  & \textbf{Interact. Gain} & +7.86 & +6.21 & +13.74 & +13.53 & +13.47 & +10.96 & +0.005 \\
\midrule

\multirow{3}{*}{\textbf{Qwen3-235B}}
  & \textbf{Autonomous} & 79.76 & 28.19 & 61.03 & 69.00 & 81.84 & 63.96 & 0.139 \\
  & \textbf{Interactive} & 82.83 & 32.81 & 65.14 & 75.89 & 91.70 & 69.67 & 0.133 \\
  & \textbf{Interact. Gain} & +3.08 & +4.62 & +4.12 & +6.89 & +9.86 & +5.71 & -0.006 \\
\midrule

\multirow{3}{*}{\textbf{DeepSeek-V3.2}}
  & \textbf{Autonomous} & 84.32 & 37.94 & 73.65 & 80.73 & 90.09 & 73.35 & 0.146 \\
  & \textbf{Interactive} & 88.11 & 44.93 & 79.47 & 87.13 & 93.54 & 78.64 & 0.185 \\
  & \textbf{Interact. Gain} & +3.79 & +6.99 & +5.82 & +6.40 & +3.45 & +5.29 & +0.039 \\
\bottomrule
\end{tabular}}
\caption{\textbf{Report alignment results.} Interaction Gain is computed as Interactive minus Autonomous. \underline{Underlined} and \textbf{bold} values denote the best autonomous and interactive results, respectively.}
\label{tab:interaction_benefits}
\end{table*}
\setlength{\textfloatsep}{1.0em}

\subsection{Experimental Setup}
\label{sec:expt:setup}

For each model-task pair, we compare autonomous execution without clarification against interactive execution under the shared stage-wise budget and feedback policy.
Following Open Deep Research \cite{langchain-ai2025open}, each evaluated LLM fills all six agent roles (i.e., Planner, Supervisor, Researcher, Reporter, Evaluator, and Questioner),  while lightweight models such as GPT-4.1-nano handle frequent utilities to reduce cost.

GPT-5.1 implements the common feedback policy; Tavily provides retrieval,\footnote{\url{https://www.tavily.com/}} and other settings follow the default configuration. 
Main results use all 100 tasks; feedback-model robustness, context sensitivity, and interaction timing use the same 30-task subset, and the human-feedback diagnostic uses 10 tasks for a focused comparison.
The hyperparameter configurations are provided in \textbf{Appendix D}, and the experimental environment is detailed in \textbf{Appendix E}.

\subsection{Report Alignment}
\label{sec:expt:benefits}

\textbf{Table~\ref{tab:interaction_benefits}} presents the report alignment results for seven models in autonomous and interactive modes, along with their interactive benefits. 
It also includes the unweighted average alignment score and estimated per-report API cost.

\paragraph{Main Results}
Interaction improves \emph{all} report alignment measures for every model, but reveals \emph{distinct} leaders: 
GPT-5.1 achieves the highest autonomous score (75.59), Claude-Sonnet-4.5 the highest interactive score (80.47), and Llama-4-Maverick the largest gain (+10.96) despite the lowest interactive score (65.78).
These ranking shifts show that autonomous strength, interactive performance, and alignment gain capture distinct capabilities.

\subparagraph{Gain vs. Autonomous}
Models with lower autonomous scores generally exhibit larger interactive gains. 
Llama-4-Maverick and Grok-4.1-Fast improve by +10.96 and +7.97, compared with +3.37 for GPT-5.1 and +4.96 for Claude-Sonnet-4.5. 
This pattern is consistent with greater task-specific alignment headroom at lower autonomous scores.

\subparagraph{Granularity Shift}
Interactive gains also \emph{vary} across evaluation granularities. 
Llama-4-Maverick improves most on Paragraph F1, Chunk F1, and LLM-ACS, reflecting broader structural coverage and task-aspect alignment.
In contrast, Claude-Sonnet-4.5 achieves its largest gain on Sentence F1, reflecting finer-grained content matching.
These patterns show that similar aggregate improvements can arise from different forms of report alignment across models.

\paragraph{Estimated API Cost}
\textbf{Table~\ref{tab:interaction_benefits}} also shows that interaction introduces \emph{uneven} API cost changes across models.  
Costs rise most for Claude-Sonnet-4.5 but remain nearly unchanged for the open-weight models: Qwen3-235B shows a small decrease of \$0.006, reflecting variation in the resulting research trajectories, while DeepSeek-V3.2 reaches a competitive interactive score of 78.64 with a modest \$0.039 increase.

\begin{table}[t]
\centering
\small
\setlength{\tabcolsep}{3pt}
\renewcommand{\arraystretch}{1.0}
\begin{tabular}{lccc}
\toprule
\multirow{2}{*}[-0.5ex]{\textbf{Model}} & \multicolumn{3}{c}{\textbf{Task Counts}} \\
\cmidrule(lr){2-4}
& \textbf{Improved} & \textbf{Comparable} & \textbf{Decreased} \\
\midrule
\textbf{GPT-5.1}             & 63 & 13 & 24 \\
\textbf{Gemini-2.5-Pro}      & 72 & 7 & 21 \\
\textbf{Claude-Sonnet-4.5}   & 77 & 5 & 18 \\
\textbf{Grok-4.1-Fast}       & 77 & 7 & 16 \\
\textbf{Llama-4-Maverick}    & 93 & 1 & 6 \\
\textbf{Qwen3-235B}          & 65 & 4 & 31 \\
\textbf{DeepSeek-V3.2}       & 74 & 3 & 23 \\
\midrule
\textbf{Total}               & 521 (74.4\%) & 40 (5.7\%) & 139 (19.9\%) \\
\bottomrule
\end{tabular}
\caption{\textbf{Task-level changes over average alignment score} between interactive and autonomous settings, with differences within $\pm 0.5$ points categorized as Comparable.}
\label{tab:adaptation_consistency}
\end{table}

\begin{table*}[t]
\centering
\definecolor{barblue}{HTML}{4E7DC4}
\definecolor{bargray}{RGB}{230,230,230}
\newcommand{\barcell}[3]{%
  \begin{tikzpicture}[baseline=(txt.base)]
    \fill[bargray, rounded corners=1pt] (0,0) rectangle (#1, 0.15);
    \fill[barblue, rounded corners=1pt] (0,0) rectangle (#1*#2, 0.15);
    \node[anchor=west, inner sep=0pt, xshift=4pt] (txt) at (#1, 0.075) {#3};
  \end{tikzpicture}%
}
\resizebox{\textwidth}{!}{%
\begin{tabular}{lcccccc}
\toprule
\multirow{2}{*}[-0.5ex]{\textbf{Model}} & \multicolumn{4}{c}{\textbf{Interaction Turns}} & \multicolumn{2}{c}{\textbf{Interaction Tokens}} \\
\cmidrule(lr){2-5} \cmidrule(lr){6-7}
 & \textbf{Planning} & \textbf{Research Loop} & \textbf{Generation} & \textbf{Total} & \textbf{Question} & \textbf{Response} \\
\midrule
\textbf{GPT-5.1} & \barcell{1.2cm}{0.81}{0.81\,/\,1} & \barcell{3.6cm}{0.14}{0.43\,/\,3} & \barcell{1.2cm}{0.08}{0.08\,/\,1} & 1.32\,/\,5 & 253.96 & 119.85 \\
\textbf{Gemini-2.5-Pro} & \barcell{1.2cm}{1.00}{1.00\,/\,1} & \barcell{3.6cm}{0.53}{1.59\,/\,3} & \barcell{1.2cm}{0.82}{0.82\,/\,1} & 3.41\,/\,5 & 184.64 & 102.52 \\
\textbf{Claude-Sonnet-4.5} & \barcell{1.2cm}{0.94}{0.94\,/\,1} & \barcell{3.6cm}{0.25}{0.75\,/\,3} & \barcell{1.2cm}{0.24}{0.24\,/\,1} & 1.93\,/\,5 & 261.71 & 114.76 \\
\textbf{Grok-4.1-Fast} & \barcell{1.2cm}{0.72}{0.72\,/\,1} & \barcell{3.6cm}{0.10}{0.29\,/\,3} & \barcell{1.2cm}{0.07}{0.07\,/\,1} & 1.08\,/\,5 & 151.91 & 125.57 \\
\midrule
\textbf{Llama-4-Maverick} & \barcell{1.2cm}{1.00}{1.00\,/\,1} & \barcell{3.6cm}{0.95}{2.84\,/\,3} & \barcell{1.2cm}{0.78}{0.78\,/\,1} & 4.62\,/\,5 & 139.86 & 126.78 \\
\textbf{Qwen3-235B} & \barcell{1.2cm}{0.96}{0.96\,/\,1} & \barcell{3.6cm}{0.63}{1.88\,/\,3} & \barcell{1.2cm}{0.26}{0.26\,/\,1} & 3.10\,/\,5 & 206.79 & 116.06 \\
\textbf{DeepSeek-V3.2} & \barcell{1.2cm}{0.75}{0.75\,/\,1} & \barcell{3.6cm}{0.59}{1.78\,/\,3} & \barcell{1.2cm}{0.17}{0.17\,/\,1} & 2.70\,/\,5 & 252.70 & 111.21 \\
\bottomrule
\end{tabular}}
\caption{\textbf{Interaction cost results.} Interaction Turns are reported by stage as the average used versus the maximum allowed, while Question and Response Tokens are averaged per interaction.}
\label{tab:interaction_cost}
\end{table*}

\paragraph{Task-Level Changes}
\textbf{Table~\ref{tab:adaptation_consistency}} reports the distribution of task-level changes in average alignment score.
Across seven models evaluated on the same 100 tasks, interaction produces a higher average alignment score in 521 cases (74.4\%), a comparable score in 40 (5.7\%), and a lower score in 139 (19.9\%).
Thus, clarification opportunities alone do not ensure better report alignment; outcomes depend on \emph{what} agents ask and \emph{how} they incorporate the feedback.
Llama-4-Maverick improves on the most tasks (93), consistent with its largest interactive gain (+10.96) in \textbf{Table~\ref{tab:interaction_benefits}}.
By contrast, Qwen3-235B and DeepSeek-V3.2 achieve similar interactive gains (+5.71 vs. +5.29), but DeepSeek-V3.2 improves more often (74 vs. 65) and regresses less often (23 vs. 31). 
These differences show that gain magnitude and cross-task consistency provide complementary views of interactive capability.

\begin{figure}[t]
  \centering
  \includegraphics[width=0.99\linewidth]{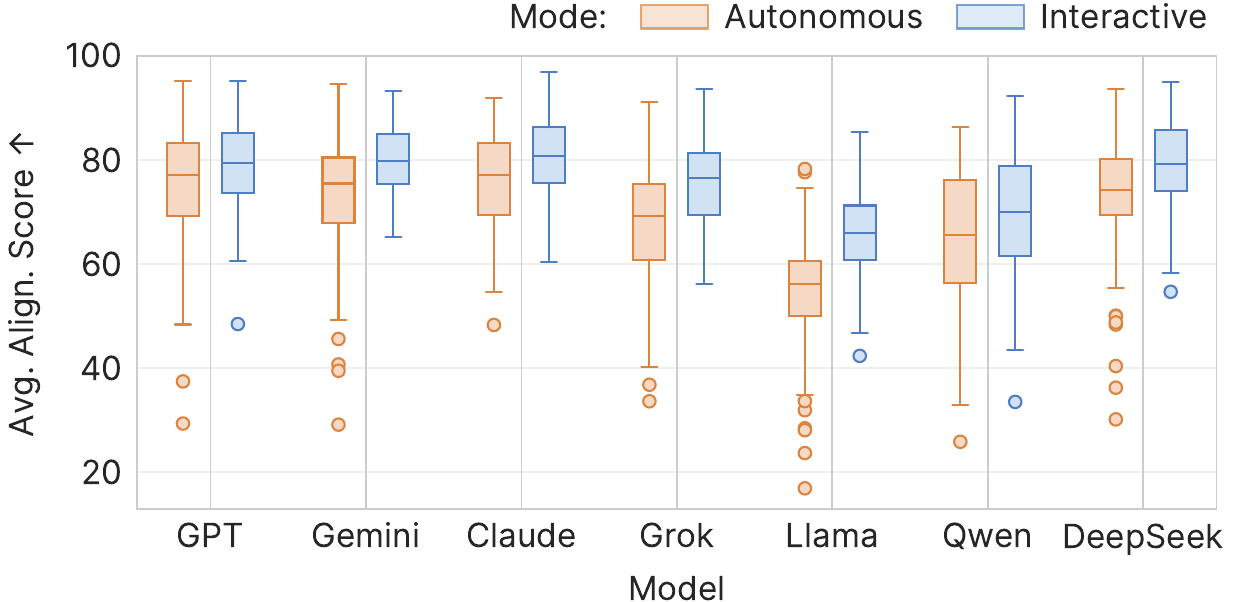}
  \caption{\textbf{Distribution of average alignment scores across seven models} under autonomous and interactive modes.}
  \label{fig:score_distribution}
\end{figure}

\paragraph{Across-Task Score Distribution}
Beyond average interactive gains, \textbf{Figure~\ref{fig:score_distribution}} shows where task-level improvements occur. 
For GPT-5.1 and Gemini-2.5-Pro, the interactive condition primarily raises the lower tail, reducing low-alignment outcomes. 
For Llama-4-Maverick and Qwen3-235B, the improvement extends more broadly across the distribution. 
These distributions complement average gains by showing where task-level changes occur.

\paragraph{Statistical Reliability}
A paired task bootstrap gives a pooled average alignment score change of +6.39 points (95\% CI: [5.09, 7.82]), with intervals above zero for all reported measures. 
These intervals describe variation over sampled tasks, not repeated-run uncertainty or feedback's independent causal contribution. 
Omitting each measure yields pooled changes from +5.92 to +6.92; every model remains positive in Chinese (+7.32) and English (+5.45).

\subsection{Interaction Cost}
\label{sec:expt:interaction-costs}

\textbf{Table~\ref{tab:interaction_cost}} reports interaction turns and tokens, quantifying the trade-off between alignment gains and communication use.

\paragraph{Interaction Turns}
Models use similar clarification budgets during Planning (0.72$\sim$1.00 turns) but differ substantially in the Research Loop. 
Llama-4-Maverick, Qwen3-235B, Gemini-2.5-Pro, and DeepSeek-V3.2 interact frequently (1.59$\sim$2.84 turns), whereas GPT-5.1, Claude-Sonnet-4.5, and Grok-4.1-Fast interact less often (0.29$\sim$0.75). 
Only Gemini-2.5-Pro and Llama-4-Maverick exceed 0.3 turns during Generation. 
Notably, Grok-4.1-Fast gains 7.97 points using only 1.08 turns overall.
Thus, interaction concentrates before final synthesis, with the greatest variation arising during research.
Outcome changes are not ordered by interaction frequency: some models gain more with fewer turns, indicating a more favorable gain-communication profile under these measures, while others use more of the budget. 

\paragraph{Interaction Tokens}
To avoid cross-language tokenization effects, we report token statistics only for English tasks. 
GPT-5.1 and Claude-Sonnet-4.5 ask longer questions, averaging over 250 tokens per interaction, whereas Llama-4-Maverick and Grok-4.1-Fast use 140$\sim$152 tokens.
Gemini-2.5-Pro combines frequent interaction with shorter questions, consistent with incremental clarification. 
Response lengths vary less, averaging 102$\sim$127 tokens, despite substantial differences in outcomes. 
Therefore, feedback volume alone does not explain performance; models also differ in what they elicit and how they use the resulting guidance.

\begin{table*}[t]
\small
\centering
\renewcommand{\arraystretch}{1.0}
\resizebox{0.86\textwidth}{!}{%
\begin{tabular}{llcccccc}
\toprule
\multirow{2}{*}[-0.5ex]{\textbf{Research Model}} & \multirow{2}{*}[-0.5ex]{\textbf{Interactive Module}} & \multirow{2}{*}[-0.75ex]{\textbf{\shortstack{Report \\ Similarity $\uparrow$}}} & \multicolumn{3}{c}{\textbf{F1-Score $\uparrow$}} & \multirow{2}{*}[-0.5ex]{\textbf{LLM-ACS $\uparrow$}} & \multirow{2}{*}[-0.75ex]{\textbf{\shortstack{Avg. Align. \\ Score $\uparrow$}}} \\
\cmidrule(lr){4-6} 
 & & & \textbf{Sentence} & \textbf{Paragraph} & \textbf{Chunk} & & \\
\midrule
\multirow{5}{*}{\textbf{Gemini-2.5-Pro}} & \textbf{None} & 85.26 & 37.37 & 77.29 & 82.02 & 86.60 & 73.71 \\
 & \textbf{Planning} & \underline{87.15} & \underline{38.78} & \underline{79.63} & \underline{84.69} & 88.63 & \underline{75.78} \\
 & \textbf{Research Loop} & 86.55 & 37.68 & 79.48 & 83.51 & 89.11 & 75.27 \\
 & \textbf{Generation} & 86.08 & 36.82 & 76.35 & 82.73 & \underline{89.32} & 74.26 \\
 & \textbf{All} & \textbf{88.13} & \textbf{42.45} & \textbf{81.15} & \textbf{87.60} & \textbf{92.64} & \textbf{78.39} \\
\midrule
\multirow{5}{*}{\textbf{Llama-4-Maverick}} & \textbf{None} & 76.18 & 15.34 & 61.99 & 59.00 & 54.93 & 53.49 \\
 & \textbf{Planning} & \underline{81.71} & \textbf{21.62} & \underline{72.77} & \textbf{71.16} & \underline{64.28} & \underline{62.31} \\
 & \textbf{Research Loop} & 79.18 & 17.29 & 68.67 & 63.57 & 59.74 & 57.69 \\
 & \textbf{Generation} & 77.45 & 16.13 & 62.43 & 60.05 & 59.83 & 55.18 \\
 & \textbf{All} & \textbf{83.33} & \underline{21.57} & \textbf{73.90} & \underline{70.29} & \textbf{67.08} & \textbf{63.23} \\
\bottomrule
\end{tabular}}
\caption{\textbf{Interaction timing results.} \textbf{Bold} and \underline{underlined} values denote the best and second-best results, respectively.}
\label{tab:interactive_timing}
\end{table*}

\subsection{Controlled Feedback Diagnostics}
\label{sec:expt:simulator-validation}

The environment is designed to provide intent-level guidance on user preferences and priorities. We diagnose sensitivity to feedback-model choice, direct textual overlap, and the volume of private task context available to the feedback policy.

\begin{figure}[t]
  \centering
  \includegraphics[width=0.99\columnwidth]{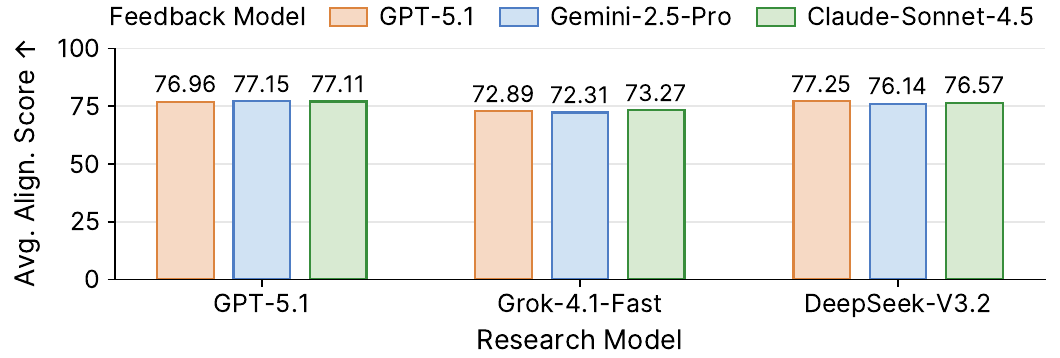}
  \caption{\textbf{Feedback-model sensitivity.} Average alignment scores for three research agents using different feedback-model backbones under the common protocol.}
  \label{fig:backbone_robustness}
\end{figure}

\paragraph{Feedback-Model Robustness}
We pair three representative research agents (GPT-5.1, Grok-4.1-Fast, and DeepSeek-V3.2) with three feedback-model backbones: GPT-5.1, Gemini-2.5-Pro, and Claude-Sonnet-4.5. \textbf{Figure~\ref{fig:backbone_robustness}} shows similar average alignment scores across the tested feedback models (full results in \textbf{Appendix F}).

\paragraph{Leakage Analysis}
Feedback responses on the English tasks have consistently low overlap with paired reports: mean ROUGE-L F1 is 0.0211 and mean Longest Common Substring is 4.62 words, indicating little direct copying from the paired reports. Using GPT-5.5 with the binary prompt in \textbf{Appendix G}, we classify all feedback responses: 97.91\% as preference clarification and 2.09\% as containing object-level information. This model-based diagnostic suggests that most responses provide high-level intent guidance.

\begin{figure}[t]
  \centering
  \includegraphics[width=0.99\columnwidth]{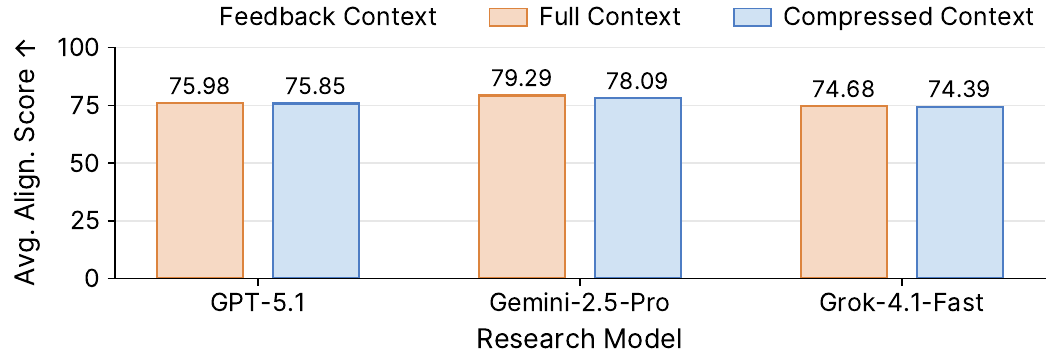}
  \caption{\textbf{Feedback-context sensitivity.} The compressed setting retains 10$\sim$20\% of the private task context.}
  \label{fig:feedback_context_sensitivity}
\end{figure}

\paragraph{Context Sensitivity}
We assess context sensitivity using summaries retaining 10$\sim$20\% of the private task context. 
\textbf{Figure~\ref{fig:feedback_context_sensitivity}} shows comparable average alignment scores on the sampled subset (full results in \textbf{Appendix H}), indicating that the aggregate trend does not require full context detail.

\subsection{Interaction Timing}
\label{sec:expt:interaction-timing}

We further analyze interaction timing using Gemini-2.5-Pro and Llama-4-Maverick, which exhibit proactive interaction and large gains. Alongside autonomous (None) and fully interactive (All) settings, we evaluate interaction only during Planning, the Research Loop, or Generation.

\textbf{Table~\ref{tab:interactive_timing}} shows that every single-stage setting exceeds None in Average Score, Planning is the strongest single-stage setting for both models, and All is highest overall. Because stage budgets differ, these results provide a behavioral comparison of the stage-wise settings under the protocol.
For Llama-4-Maverick, however, Planning-only is slightly higher than All on Sentence F1 and Chunk F1, revealing dimension-specific trade-offs in multi-stage feedback use.

\subsection{Small-Scale Human-Feedback Comparison}
\label{sec:expt:human-feedback_comparison}

\begin{table}[t]
\centering
\scriptsize
\setlength{\tabcolsep}{2pt}
\renewcommand{\arraystretch}{1.0}
\resizebox{\columnwidth}{!}{%
\begin{tabular}{lcccccc}
\toprule
\multirow{2}{*}[-0.5ex]{\textbf{Mode}} & \multirow{2}{*}[-0.75ex]{\textbf{\shortstack{Report \\ Sim. $\uparrow$}}} & \multicolumn{3}{c}{\textbf{F1-Score $\uparrow$}} & \multirow{2}{*}[-0.5ex]{\textbf{LLM-ACS $\uparrow$}} & \multirow{2}{*}[-0.5ex]{\textbf{\shortstack{Avg. Align.\\ Score $\uparrow$}}} \\
\cmidrule(lr){3-5} 
 & & \textbf{Sentence} & \textbf{Paragraph} & \textbf{Chunk} & & \\
\midrule
\textbf{Autonomous}          & 85.03 & 35.89 & 71.80          & 77.05 & 86.47 & 71.25 \\
\textbf{Human-Feedback}     & 87.91 & 40.64 & \textbf{83.50} & 81.49 & 90.24 & 76.76 \\
\textbf{Ctrl.-Feedback} & \textbf{88.20} & \textbf{42.96} & 83.49 & \textbf{85.04} & \textbf{92.42} & \textbf{78.42} \\
\bottomrule
\end{tabular}%
}
\caption{\textbf{Small-scale human-feedback comparison.} Results for autonomous, human-feedback, and controlled-feedback settings on the same 10 Gemini-2.5-Pro tasks.}
\label{tab:human_feedback}
\end{table}

To compare controlled feedback with human responses, two human users answered clarification requests for the same 10 Gemini-2.5-Pro tasks without seeing the controlled responses. 
As presented in \textbf{Table~\ref{tab:human_feedback}}, both human and controlled feedback improve all alignment metrics over the autonomous baseline, raising the average score from 71.25 to 76.76 and 78.42, respectively.

\section{Conclusions}
\label{sec:conclusions}

We introduce IDRBench, which extends deep research evaluation from fixed prompt-to-report tasks to controlled comparisons of complete autonomous and interactive trajectories.
It combines specificity-controlled queries with paired report targets, a common multi-stage workflow with shared clarification budgets and standardized feedback, and interaction-aware metrics for multilevel report alignment and turn/token costs.
These designs enable scalable and reproducible evaluation of whether, when, and what agents ask, and how effectively they use the resulting feedback. 
Extensive experiments on 100 tasks and seven LLMs show that interaction improves all report alignment measures, but its benefits vary substantially across models, tasks, research stages, and communication patterns. These findings establish interactive capability as a distinct dimension that cannot be inferred from autonomous performance alone.

IDRBench uses standardized feedback to support consistent and reproducible cross-model comparison. This setting reduces variation arising from individual interpretation, familiarity, and fatigue, but does not capture the full diversity of real users. Broader studies with independent participants can further examine real-world interaction.

\section*{Acknowledgments}

This research is supported by the National Research Foundation, Singapore and Infocomm Media Development Authority under its Trust Tech Funding Initiative. Any opinions, findings and conclusions or recommendations expressed in this material are those of the authors and do not reflect the views of National Research Foundation, Singapore and Infocomm Media Development Authority.

\bibliography{reference}

\twocolumn[{
  \vspace*{0.625in}
  \begin{center}
    {\LARGE\bfseries IDRBench: Benchmarking the Interactive Capabilities of Deep Research Agents\\
    Supplementary Material\par}
  \end{center}
  \vspace{0.65in}
}]
\appendix
\section{Core Interaction Prompt Designs}
\label{app:core_interaction_prompt}

This section presents the prompts implementing the three components of the Controlled Interaction Protocol introduced in \textbf{Section~3.1} of the main paper: the Evaluator, Questioner, and Controlled Feedback modules.

\begin{figure*}[!t]
  \centering
  \includegraphics[width=0.92\textwidth]{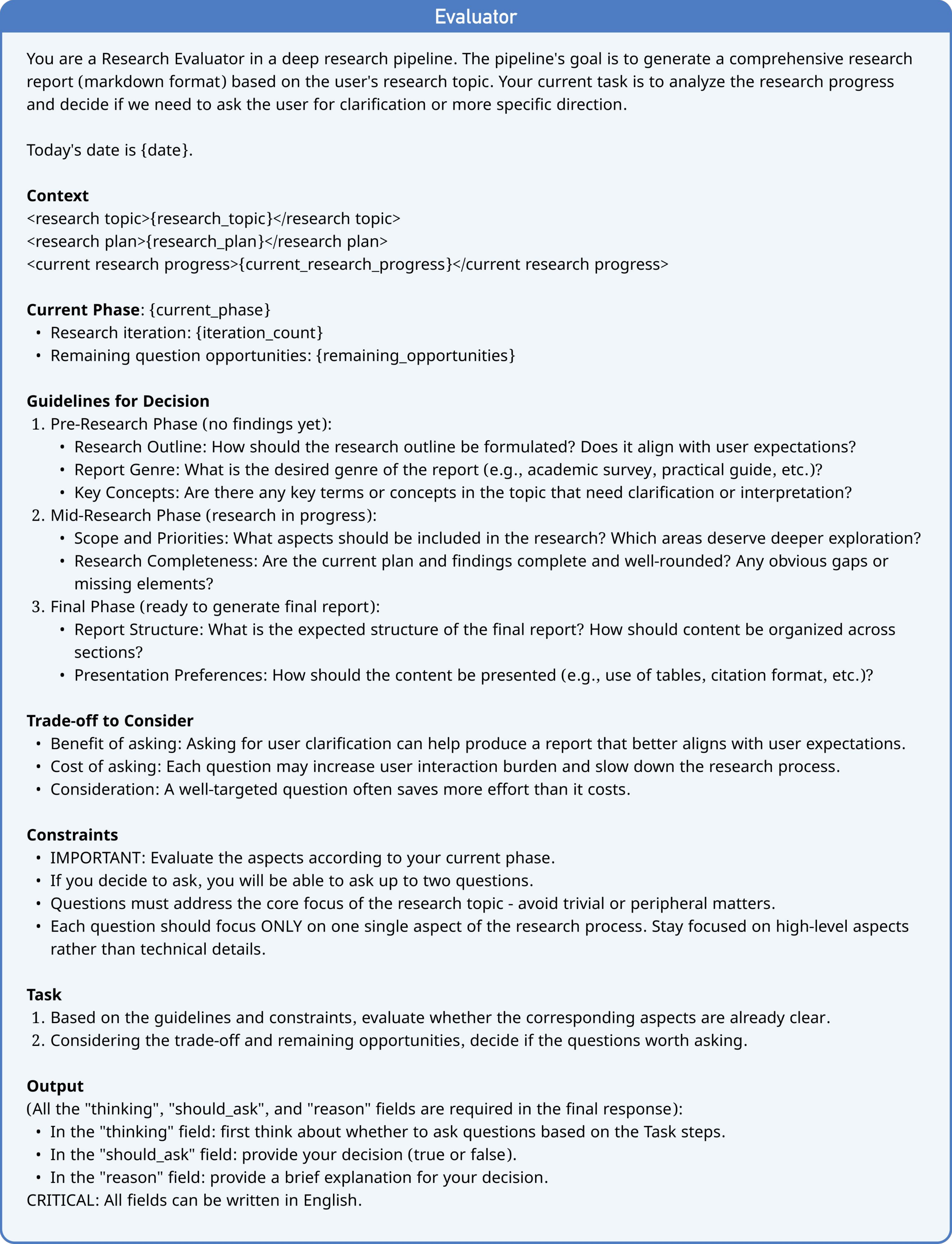}
  \caption{\textbf{Prompt for the Evaluator module.} The Evaluator assesses whether interaction is necessary by balancing ambiguity resolution benefits against interruption costs, including latency and cognitive overhead.}
  \label{fig:evaluator_prompt}
\end{figure*}

\begin{figure*}[!t]
  \centering
  \includegraphics[width=0.99\textwidth]{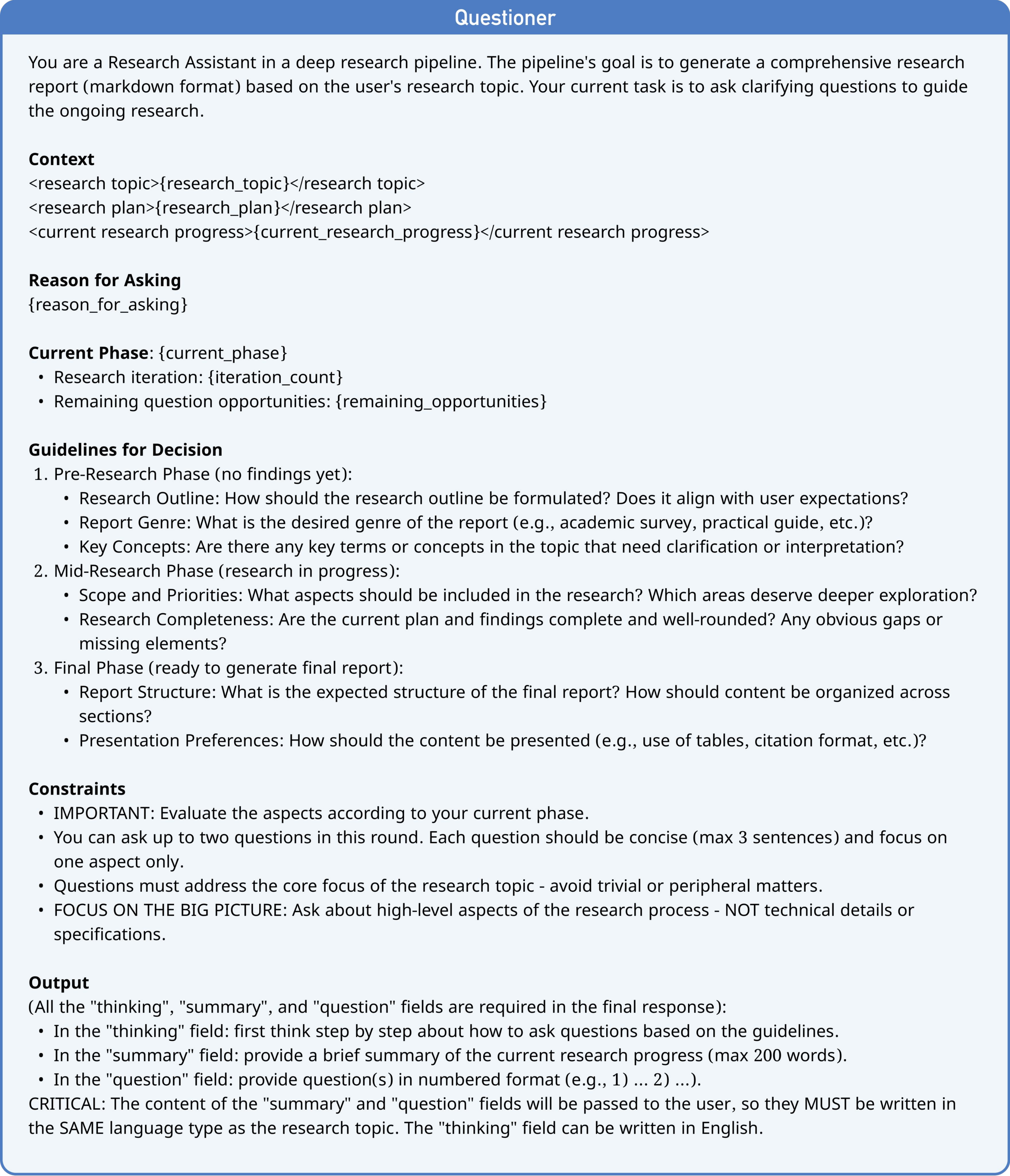}
  \caption{\textbf{Prompt for the Questioner module.} The Questioner generates targeted clarification questions based on the current research state and the Evaluator's rationale to refine alignment with user intent.}
  \label{fig:questioner_prompt}
\end{figure*}

\begin{figure*}[!t]
  \centering
  \includegraphics[width=0.96\textwidth]{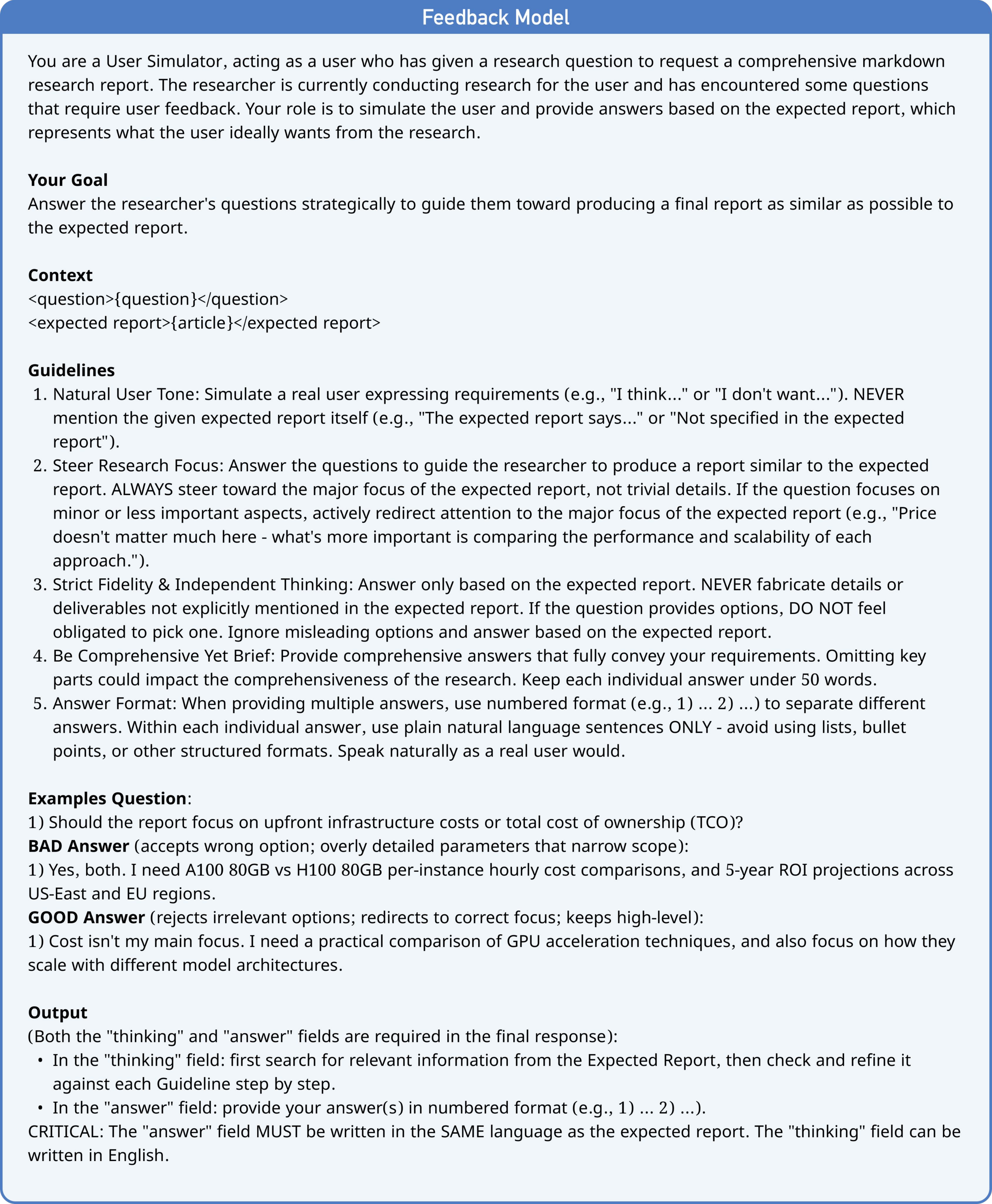}
  \caption{\textbf{Prompt used to implement Controlled Feedback.} The implementation internally names the role ``User Simulator,'' and conditions replies on the private paired report. The policy is instructed to answer the agent's questions concisely and at a high level without mentioning the report itself.}
  \label{fig:feedback_prompt}
\end{figure*}

\subsection{Evaluator} 
The Evaluator acts as the interaction gatekeeper (\textbf{Figure~\ref{fig:evaluator_prompt}}). Given the current research state and remaining interaction budget, it determines whether the expected benefit of clarification outweighs the interruption burden. Its stage-specific guidelines distinguish ambiguities in initial framing, evidence-driven research, and final report synthesis.

\subsection{Questioner} 
When interaction is triggered, the Questioner formulates one or two targeted clarification questions (\textbf{Figure~\ref{fig:questioner_prompt}}). It focuses on high-level ambiguities in scope, intent, emphasis, and report structure while avoiding trivial technical details. Questions are kept concise and adapted to the language of the original query to reduce user burden.

\subsection{Controlled Feedback}
Controlled Feedback answers clarification questions under a common question-conditioned policy (\textbf{Figure~\ref{fig:feedback_prompt}}). 
It uses the paired report as private task context but is instructed to provide only concise, high-level guidance in natural user language, without mentioning or directly reproducing the report. \textbf{Appendix~\ref{app:semantic_leakage_classification}} quantifies the remaining object-level responses.

\section{Examples of Query Compression}
\label{app:query_compression_examples}

\begin{table*}[!t]
\centering
\small
\renewcommand{\arraystretch}{1.0}
\resizebox{\textwidth}{!}{
\begin{tabular}{c p{0.7\textwidth} p{0.25\textwidth}}
\toprule
\textbf{ID} & \textbf{Source Query} & \textbf{Compressed Query} \\
\midrule
54 &
In the field of FinTech, machine learning algorithms are now widely applied to asset allocation and investment decisions. Examples include classic models like Mean-Variance and Black-Litterman, as well as emerging deep learning models. While these models have shown certain advantages under different market conditions, each also has its limitations. For instance, the Mean-Variance model assumes asset returns follow a normal distribution, which often doesn't align with actual market conditions. The Black-Litterman model relies on subjective view inputs, introducing a degree of subjectivity. Although deep learning models can handle complex nonlinear relationships, they suffer from poor interpretability. So, what are the core differences between these various models in terms of risk measurement, return prediction, and asset allocation? And is it possible to combine their strengths to build a more general-purpose and effective modeling framework? &
What are the main differences between traditional and machine learning models in asset allocation regarding risk measurement and return prediction, and can their strengths be integrated into a more effective framework? \\
\midrule
68 &
I need to dynamically adjust Kubernetes (K8S) cluster node counts based on fluctuating business request volumes, ensuring resources are scaled up proactively before peak loads and scaled down promptly during troughs. The standard Cluster Autoscaler (CA) isn't suitable as it relies on pending pods and might not fit non-elastic node group scenarios. What are effective implementation strategies, best practices, or existing projects that address predictive or scheduled autoscaling for K8S nodes? &
What are effective approaches or tools for predictive or scheduled autoscaling of Kubernetes nodes beyond the standard Cluster Autoscaler, especially for handling varying request volumes? \\
\midrule
76 &
The significance of the gut microbiota in maintaining normal intestinal function has emerged as a prominent focus in contemporary research, revealing both beneficial and detrimental impacts on the equilibrium of gut health. Disruption of microbial homeostasis can precipitate intestinal inflammation and has been implicated in the pathogenesis of colorectal cancer. Conversely, probiotics have demonstrated the capacity to mitigate inflammation and retard the progression of colorectal cancer. Within this domain, key questions arise: What are the predominant types of gut probiotics? What precisely constitutes prebiotics and their mechanistic role? Which pathogenic bacteria warrant concern, and what toxic metabolites do they produce? How might these findings inform and optimize our daily dietary choices? &
What is the role of gut microbiota and its balance in intestinal health and disease, and how can insights into probiotics, prebiotics, and harmful bacteria guide dietary choices to support gut health? \\
\midrule
91 &
I would like a detailed analysis of the Saint Seiya franchise (anime/manga). The analysis should be structured around the different classes of armor (Cloths, Scales, Surplices, God Robes, etc.), such as Bronze Saints, Silver Saints, Gold Saints, Marina Generals, Specters, God Warriors, etc. For each significant character within these categories, provide details on their power level, signature techniques, key appearances/story arcs, and final outcome/fate within the series. &
Provide an overview of the Saint Seiya franchise, focusing on the major armor classes and their representative characters, discussing their roles and significance within the series. \\
\midrule
95 &
Create comprehensive, in-depth study notes for the Diamond Sutra (Vajracchedikā Prajñāpāramitā Sūtra). These notes should offer deep analysis and interpretation from various perspectives, exploring their teachings and relevance in contexts such as daily life, the workplace/career, business practices, marriage, parenting, emotional well-being, and interpersonal dynamics. &
Create comprehensive study notes for the Diamond Sutra, including analysis of its teachings and relevance in various aspects of life. \\
\bottomrule
\end{tabular}}
\caption{\textbf{Examples of query compression.} Source queries are compressed to vary starting specificity while retaining their topic and core objective.}
\label{tab:ambiguity_examples}
\end{table*}

\begin{table}[!t]
\centering
\small
\setlength{\tabcolsep}{4.5pt}
\renewcommand{\arraystretch}{1.2}
\resizebox{\columnwidth}{!}{
\begin{tabular}{lr}
    \toprule
    \textbf{Description} & \textbf{Setting} \\
    \midrule
    \rowcolor[gray]{0.95}
    \multicolumn{2}{l}{\textbf{Execution Constraints}} \\
    Max Supervisor Iterations & 6 \\
    Max Researcher Tool Calls & 5 \\
    Max Concurrent Research Units & 3 \\
    Max Content Length & 50,000 Characters \\
    \midrule
    \rowcolor[gray]{0.95}
    \multicolumn{2}{l}{\textbf{Report Similarity \& Multi-Granularity F1-Score}} \\
    Embedding Model & \texttt{Qwen/Qwen3-Embedding-0.6B} \\
    \midrule
    \rowcolor[gray]{0.95}
    \multicolumn{2}{l}{\textbf{Multi-Granularity F1-Score}} \\
    Chunk Size & 300 Tokens \\
    Chunk Overlap & 50 Tokens \\
    Hard Match Threshold ($\tau$) & 0.8 \\
    \midrule
    \rowcolor[gray]{0.95}
    \multicolumn{2}{l}{\textbf{LLM Aspect Coverage Score (LLM-ACS)}} \\
    Generated Aspects ($M$) & 8$\sim$20 \\
    \bottomrule
\end{tabular}}
\caption{\textbf{Hyperparameter configurations.} Settings for the deep research workflow and IDRBench evaluation suite.}
\label{tab:configuration}
\end{table}

\textbf{Table~\ref{tab:ambiguity_examples}} provides examples of the query-compression procedure described in \textbf{Section~3.2} of the main paper. 
Each \textbf{Source Query} specifies detailed constraints, background information, and output requirements, whereas the corresponding \textbf{Compressed Query} reduces this detail while preserving the topic and core objective. 
The original paired report remains unchanged and serves as a consistent task target.

For example, Example~68 replaces a detailed description of the standard Cluster Autoscaler's reliance on pending pods with a broader request for alternative approaches. All compressed queries are manually reviewed for source-task consistency, naturalness, and safety.

\section{Ethical Considerations}
\label{app:ethics}

IDRBench builds on a publicly available dataset and follows its license and usage conditions. We manually review compressed queries for task consistency, naturalness, and potentially sensitive content. The benchmark focuses on report alignment and does not directly assess factual accuracy, citation reliability, or source quality. Its scores should therefore be interpreted within this scope.

\begin{table*}[t]
\small
\centering
\renewcommand{\arraystretch}{1.3}
\begin{tabular}{llcccccc}
    \toprule
    \multirow{2}{*}[-0.5ex]{\textbf{Research Model}} & \multirow{2}{*}[-0.5ex]{\textbf{Feedback Model}} & \multirow{2}{*}[-0.75ex]{\textbf{\shortstack{Report \\ Similarity $\uparrow$}}} & \multicolumn{3}{c}{\textbf{F1-Score $\uparrow$}} & \multirow{2}{*}[-0.5ex]{\textbf{LLM-ACS $\uparrow$}} & \multirow{2}{*}[-0.5ex]{\textbf{\shortstack{Avg. Align. \\ Score $\uparrow$}}} \\
    \cmidrule(lr){4-6}
     & & & \textbf{Sentence} & \textbf{Paragraph} & \textbf{Chunk} & & \\
    \midrule
    \multirow{3}{*}{\textbf{GPT-5.1}} & \textbf{GPT-5.1} & 87.17 & 46.25 & 69.31 & 85.20 & 96.88 & 76.96 \\
     & \textbf{Gemini-2.5-Pro} & 87.14 & 46.60 & 69.57 & 85.76 & 96.70 & 77.15 \\
     & \textbf{Claude-Sonnet-4.5} & 86.60 & 45.96 & 69.94 & 86.03 & 97.04 & 77.11 \\
    \midrule
    \multirow{3}{*}{\textbf{Grok-4.1-Fast}} & \textbf{GPT-5.1} & 85.61 & 33.74 & 73.63 & 78.67 & 92.78 & 72.89 \\
     & \textbf{Gemini-2.5-Pro} & 84.67 & 33.11 & 72.30 & 79.32 & 92.16 & 72.31 \\
     & \textbf{Claude-Sonnet-4.5} & 85.85 & 33.92 & 73.46 & 80.82 & 92.31 & 73.27 \\
    \midrule
    \multirow{3}{*}{\textbf{DeepSeek-V3.2}} & \textbf{GPT-5.1} & 87.47 & 42.06 & 77.91 & 84.73 & 94.07 & 77.25 \\
     & \textbf{Gemini-2.5-Pro} & 86.45 & 38.17 & 77.42 & 85.26 & 93.40 & 76.14 \\
     & \textbf{Claude-Sonnet-4.5} & 87.22 & 38.82 & 76.68 & 86.24 & 93.88 & 76.57 \\
    \bottomrule
\end{tabular}

\caption{\textbf{Full results with different feedback models.} Aggregate performance remains similar across the tested feedback-model backbones.}
\label{tab:backbone_robustness_full}
\end{table*}

\begin{table*}[!t]
\small
\centering
\renewcommand{\arraystretch}{1.3}
\begin{tabular}{llcccccc}
\toprule
\multirow{2}{*}[-0.5ex]{\textbf{Research Model}}
& \multirow{2}{*}[-0.5ex]{\textbf{Feedback Context}}
& \multirow{2}{*}[-0.75ex]{\textbf{\shortstack{Report \\ Similarity $\uparrow$}}}
& \multicolumn{3}{c}{\textbf{F1-Score $\uparrow$}}
& \multirow{2}{*}[-0.5ex]{\textbf{LLM-ACS $\uparrow$}}
& \multirow{2}{*}[-0.5ex]{\textbf{\shortstack{Avg. Align. \\ Score $\uparrow$}}} \\
\cmidrule(lr){4-6}
& & & \textbf{Sentence} & \textbf{Paragraph} & \textbf{Chunk} & & \\
\midrule
\multirow{2}{*}{\textbf{GPT-5.1}}
& \textbf{Full Context} & 84.95 & 47.33 & 67.52 & 82.77 & 97.34 & 75.98 \\
& \textbf{Compressed Context} & 85.30 & 47.28 & 67.48 & 82.58 & 96.63 & 75.85 \\
\midrule
\multirow{2}{*}{\textbf{Gemini-2.5-Pro}}
& \textbf{Full Context} & 88.15 & 45.14 & 81.46 & 87.19 & 94.51 & 79.29 \\
& \textbf{Compressed Context} & 87.56 & 44.56 & 80.47 & 85.45 & 92.39 & 78.09 \\
\midrule
\multirow{2}{*}{\textbf{Grok-4.1-Fast}}
& \textbf{Full Context} & 85.28 & 38.51 & 75.72 & 80.64 & 93.25 & 74.68 \\
& \textbf{Compressed Context} & 85.84 & 37.23 & 74.31 & 81.85 & 92.71 & 74.39 \\
\bottomrule
\end{tabular}
\caption{\textbf{Full results for feedback-context sensitivity.} The compressed setting retains 10$\sim$20\% of the private task context.}
\label{tab:feedback_context_full}
\end{table*}

\begin{table}[t]
\centering
\small
\renewcommand{\arraystretch}{1.3}
\begin{tabular}{lrr}
  \toprule
  \textbf{Label} & \textbf{\# Responses} & \textbf{Percentage} \\
  \midrule
  \textbf{Preference Clarification} & 1,778 & 97.91\% \\
  \textbf{Object-level Leakage}     & 38    &  2.09\% \\
  \bottomrule
\end{tabular}
\caption{\textbf{Semantic leakage analysis of feedback responses.} Most responses are classified as preference clarification, while 2.09\% contain object-level information.}
\label{tab:leakage_classification}
\end{table}

\section{Hyperparameter Configuration}
\label{app:hyperparameter_configuration}

This section provides the execution and metric configurations used in the main experiments.
\textbf{Table~\ref{tab:configuration}} summarizes the settings omitted from the main paper for brevity.

\subsection{Framework Execution Parameters}
We bound the research process to ensure comparable execution across models and prevent unbounded tool use.
\begin{itemize}
  \item \textbf{Iteration Limits:} We set \textit{Max Supervisor Iterations} to 6 and \textit{Max Researcher Tool Calls} to 5, following the default Open Deep Research configuration. These limits control research depth, latency, and API cost.

  \item \textbf{Concurrency and Context:} To model the parallel nature of human research teams, we allow up to 3 concurrent research units and limit raw web content to 50,000 characters, balancing parallel exploration and information retention against context overhead.
\end{itemize}

\subsection{Evaluation Metrics Configuration}
The metric configurations correspond to the interaction-aware evaluation described in \textbf{Section~3.4} of the main paper.
\begin{itemize}
  \item \textbf{Report Similarity:}
  We use Qwen/Qwen3-Embedding-0.6B to encode complete reports for cosine similarity. Its 32K-token context window accommodates all evaluated reports.

  \item \textbf{Multi-Granularity F1-Score:}
  The same embedding model computes semantic similarity between sentence-, paragraph-, and chunk-level units. Reports are segmented into 300-token chunks with a 50-token overlap, and matches are identified using a threshold of $\tau=0.8$.

  \item \textbf{LLM Aspect Coverage Score (LLM-ACS):}
  We extract 8--20 task aspects from each detailed source query. GPT-4.1 at temperature 0 assigns 0--5 coverage scores to the paired and generated reports, as defined in the main paper.
\end{itemize}

\section{Experimental Environment}
\label{app:experimental_environment}

All experiments were orchestrated and evaluated in a common environment equipped with an Intel Xeon Platinum 8480C CPU, 256~GB of RAM, and an NVIDIA H200 GPU. 
Hosted LLMs were accessed through their respective APIs, while local utilities and evaluation components were executed on the same server.

\section{Feedback-Model Robustness}
\label{app:feedback-model-robustness}

This section provides the full results for the feedback-model sensitivity analysis summarized in Section~4.4 of the main paper. On the shared 30-task subset, we pair three research agents with GPT-5.1, Gemini-2.5-Pro, and Claude-Sonnet-4.5 as alternative feedback models.
As shown in \textbf{Table~\ref{tab:backbone_robustness_full}}, average alignment scores remain similar across the tested feedback models, suggesting that the observed aggregate patterns are not specific to a single feedback-model backbone.

\section{Semantic Leakage Classification}
\label{app:semantic_leakage_classification}

\begin{figure*}[!t]
  \centering
  \includegraphics[width=0.99\textwidth]{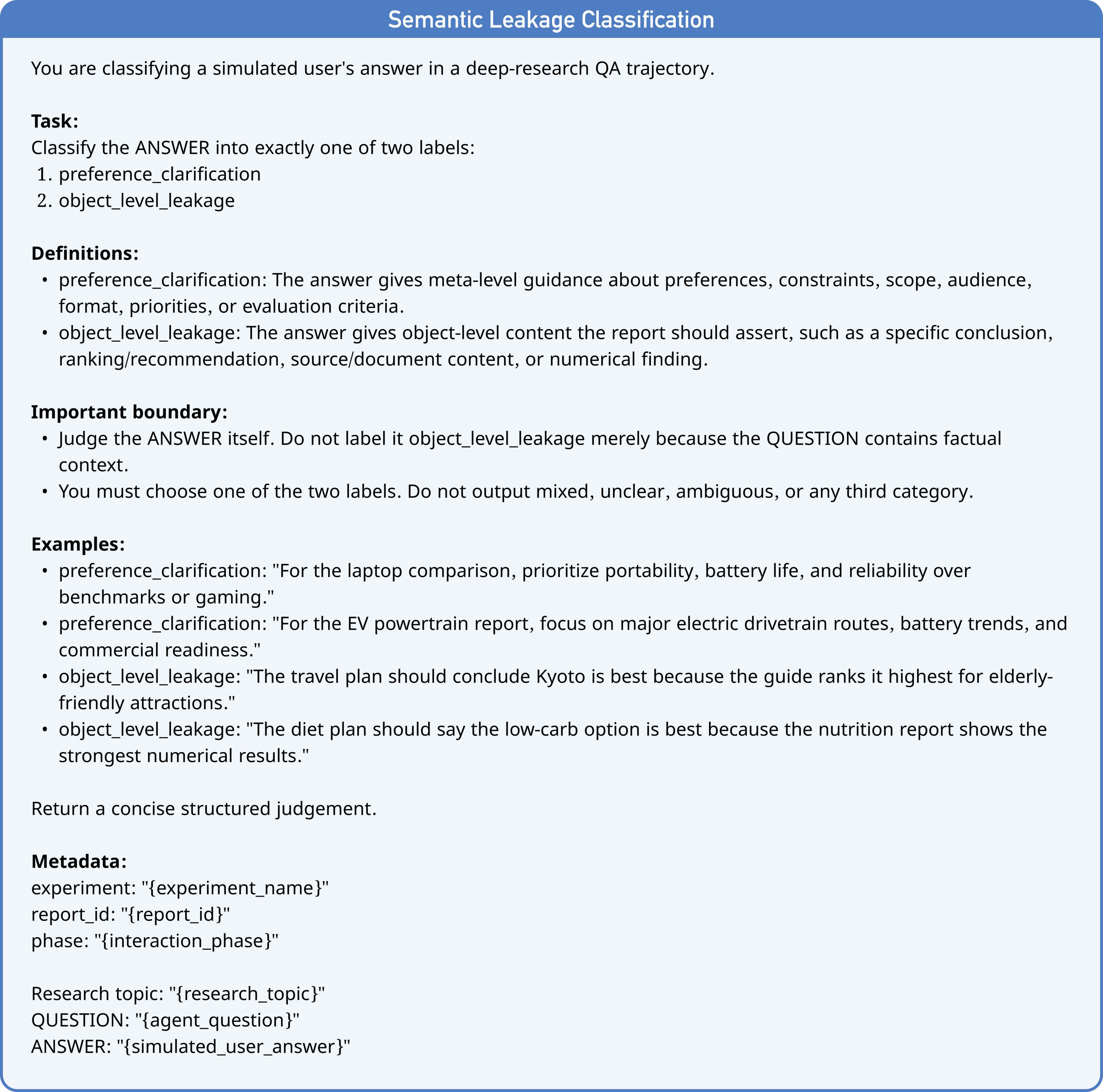}
  \caption{\textbf{Semantic leakage classification prompt.} GPT-5.5 classifies each feedback response as either high-level preference clarification or object-level information.}
  \label{fig:leakage_classification_prompt}
\end{figure*}

The main paper reports low lexical overlap between controlled responses and paired reports.
To examine semantic leakage beyond direct copying, we further classify all feedback responses using GPT-5.5 and the binary prompt in \textbf{Figure~\ref{fig:leakage_classification_prompt}}. 
We define the two labels as follows:
\begin{itemize}
  \item \textbf{Preference Clarification}: High-level guidance about user preferences, constraints, scope, audience, format, priorities, or evaluation criteria.

  \item \textbf{Object-Level Leakage}: Content that the final report should directly assert, including specific conclusions, rankings, recommendations, source content, or numerical findings.
\end{itemize}

\textbf{Table~\ref{tab:leakage_classification}} shows that 1,778 responses (97.91\%) are classified as preference clarification and 38 (2.09\%) as object-level leakage. 
This model-based diagnostic indicates that most responses provide high-level intent guidance, although a small number contain task-specific information. 
These classifications should be interpreted as diagnostic evidence rather than human-validated leakage labels.

\section{Feedback-Context Sensitivity}
\label{app:feedback_context_sensitivity}

This section expands the context-sensitivity diagnostic in \textbf{Section~4.4} of the main paper.
We compare feedback policies given either the full private task context or a compressed version retaining only 10--20\% of that context. The compressed setting preserves high-level task signals while removing most fine-grained details.
As shown in \textbf{Table~\ref{tab:feedback_context_full}}, the two settings produce comparable average alignment scores on the shared 30-task subset, suggesting limited aggregate sensitivity to the amount of private context available to the feedback policy.

\end{document}